%% file: nips13_at.tex
\documentclass{article} 
\usepackage{nips13submit_e,times}
\usepackage{hyperref}
\usepackage{url}
\usepackage{verbatim}
\usepackage{graphicx}
\usepackage{caption}
\usepackage{subcaption}

\usepackage{amsthm}

\theoremstyle{definition}

\title{Fast Gradient-Based Inference \\with Continuous Latent Variable Models\\in Auxiliary Form}

\author{
Diederik P. Kingma \\
Intelligent Autonomous Systems Group \\
Universiteit van Amsterdam \\
\texttt{dpkingma@gmail.com} \\
}

\input{commands.tex}

\nipsfinalcopy 

\begin{document}

\maketitle

\begin{abstract}
We propose a technique for increasing the efficiency of gradient-based inference and learning in Bayesian networks with multiple layers of continuous latent variables.
We show that, in many cases, it is possible to express such models in an auxiliary form, where continuous latent variables are conditionally deterministic given their parents and a set of independent auxiliary variables. Variables of models in this auxiliary form have much larger Markov blankets, leading to significant speedups in gradient-based inference, e.g. rapid mixing Hybrid Monte Carlo and efficient gradient-based optimization. The relative efficiency is confirmed in experiments.
\end{abstract}

\section{Introduction and related work}

\emph{Bayesian networks} (also called \emph{belief networks}) are probabilistic graphical models where a set of random variables and their conditional dependencies are described by a directed acyclic graph (DAG). Many supervised and unsupervised models can be considered as special cases of Bayesian networks.

In this paper we focus on the problem of efficient inference in Bayesian networks with multiple layers of continuous latent variables, where exact inference is intractable (e.g. the conditional dependencies between variables are nonlinear) but the joint distribution is differentiable. 
Algorithms for approximate inference in dynamic Bayesian networks can be roughly divided into two categories: sampling approaches and parametric approaches. Parametric approaches include Belief Propagation~\cite{pearl1982reverend} or the more recent Expectation Propagation (EP)~\cite{minka2001expectation}). When it is not reasonable or possible to make assumptions about the posterior (which is often the case), one needs to resort to sampling approaches such as Markov Chain Monte Carlo (MCMC)~\cite{neal1993probabilistic}. 
In high-dimensional spaces, gradient-based samplers such as Hybrid Monte Carlo~\cite{duane1987hmc} are known for their relatively fast mixing properties.
When just interested in finding a mode of the posterior, vanilla gradient-based optimization methods can be used.

Our method uses the concept of conditionally deterministic variables, which have been earlier used in e.g.~\cite{cobb2005nonlinear}, but not as a tool for efficient inference in general Bayesian networks with continuous variables.

In section~\ref{sec:auxform} we propose a method for transforming a Bayesian network with continuous latent variables to an (equivalent) auxiliary model, where the continuous latent variables $\bb{Z}$ are replaced by latent variables $\btZ$ that are conditonally deterministic given auxiliary variables $\bb{E}$. We show that this auxiliary model is equivalent to the original model when the auxiliary variables $\bb{E}$ are integrated out. However, it is also possible to integrate out the $\btZ$, resulting in an auxiliary joint PDF over $\bb{X}$ and $\bb{E}$. The main advantage is that inference is much faster in this form.
This can be explained from the observation that the Markov blankets of variables in the auxiliary form are larger than the variables in the original form; consequently, gradients of the auxiliary PDF take into account information from more variables, while the computational complexity for each update is equal to the original form.
In section~\ref{sec:experiments} we show some empirical results that confirm our theoretical results. Gains are especially noticable in models with many interconnected latents variables, such as Dynamic Bayesian networks~\cite{murphy2002dynamic}.

The method is applicable to conditional distributions for which it is possible to design an auxiliary form (see section~\ref{validauxforms}). This includes distributions for which a differentiable inverse CDF, or differentiable approximations, exists.








\section{Background}

\paragraph{Notes about notation.} 
We write $\pT(X|Y)$ and $\pT(X)$ to denote distributions with parameter $\bT$, i.e. $\pT(.)$ is shorthand for $p(.;\bT)$. We write $\pT(\bx|\by)$ and $\pT(\bx)$ to denote $\pT(X=\bx|Y=\by)$ and $\pT(X=\bx)$, the corresponding (conditional) probability density or mass functions (PDFs or PMFs). In some situations $\bT$ is treated as a random variable. Also note $\bT$ is treated as a global vector, and each distribution typically only uses a small subset of $\bT$'s elements. Variables are capitalized, sets of variables are capitalized and bold, vectors are written in bold and lower case.

\subsection{Bayesian networks}\label{sec:bayesnets}

A Bayesian network models a set of random variables $\bb{V} = \{V_j: j \in \{1 \dots N\}$ and their conditional dependencies as a directed acyclic graph, where each variable corresponds to a vertex and each edge to a conditional dependency. Let the distribution of each variable $V_j$ be $\pT(V_j|\bb{Pa}_j)$, where we condition on $V_j$'s (possibly empty) set of parents $\bb{Pa}_j$, and $\bT$ is a global parameter vector. The joint distribution over $\bb{V}$ can be computed as follows, using the factorization property of Bayesian networks:
\begin{align}\label{eq:bnjoint}
\pT(\bb{V}) = \pT(V_1, \dots, V_N) = \prod_{j=1}^N \pT(V_j|\bb{Pa}_j)
\end{align}
The joint PDF of Bayesian networks can be represented as a \emph{factor graph} where each factor $f_{V_j}(.)$ corresponds to an individual conditional PDF: $f_{V_j}(\bv_j; \bpa_j, \bT) = \pT(\bv_j|\bpa_j)$ where $\bv_j$ is a value of $V_j$, $\bpa_j$ is a value of $\bb{Pa}_j$ and $\pT(\bv_j|\bpa_j)$ is the PDF or PMF of the corresponding conditional distribution $\pT(V_j|\bb{Pa}_j)$ \,\footnote{For brevity we sometimes treat sets of random variables like $\bb{Pa}_j$ as a random variables and $\bpa_j$ as their instantiated vectors}.
The \emph{Markov blanket} of some variable $V_j$ can be described is the set of all variables that appear as arguments of factors in which $V_j$ also appears as an argument.

We use \emph{conditionally deterministic variables}. The value of such a variable $V_j$ with parents $\bb{Pa}_j$ is deterministically computed with a (possibly nonlinear) function $g_j(.)$ from the parents and the parameters in the Bayesian network:
$V_j = g_j(\bb{Pa}_j, \bT)$.
The PDF of a conditionally deterministic variable is a Dirac delta function, which we define as a Gaussian PDF $\mathcal{N}(.;\mu,\sigma)$ with infinitesimal $\sigma$:
\begin{align}
\pT(\bv_j|\bb{pa}_j) = \lim_{\sigma \to 0} \mathcal{N}(\bv_j; g_j(\bpa_j, \bT), \sigma)
\label{diracPDF}
\end{align}
which equals $+\infty$ when $\bv_j = g_j(\bpa_j, \bT)$ and equals 0 everywhere else, and $\int_{\bv_j} \pT(\bv_j|\bb{pa}_j) = 1$. 

\subsection{Learning problem under consideration}\label{problem}
Let $\bx = \{\bx^{(i)}: i = 1 \dots M\}$ be some i.i.d. dataset with $M$ datapoints where $\bx$ is a vector with all data concatenated. Likewise, $\bz = \{\bz^{(i)}: i = 1 \dots M\}$ is a value of all latent variables variables for all datapoints. The joint PDF over data data and latent variables is $\pT(\bx, \bz) = \prod_{i=1}^M \pT(\bx^{i}|\bz^{i})$.
We focus on the problem of learning the parameters of Bayesian networks with continuous latent variables where the data likelihood $\pT(\bx) = \int_\bz \pT(\bx, \bz) \,d\bz$ is intractable to compute or differentiate (which is true in general). We also focus on the case where the joint distribution $\pT(\bx, \bz)$ over all variables is (at least) once differentiable, so it is possible to efficiently compute first-order partial derivatives of the joint distribution w.r.t. variables and parameters.

During inference we often treat the parameters $\bT$ as a random variable where we let $p(.|\bT) = \pT(.)$. In this case, two well-known modes of learning can be distinghuised: (1) maximum \emph{a posteriori} (MAP) inference where we are interested in a value of $\bT$ that (approximately) maximizes $p(\bT | \bx) \propto p(\bx | \bT) p(\bT)$, and (2) full Bayesian inference where we are interested in finding (or approximating) the full posterior distribution $p(\bT | \bx)$.

\subsection{Gradient-based learning algorithms}\label{algorithms}

For the learning problem outlined in section \ref{problem}, we summarize some well-known gradient-based learning approaches using approximate inference.

\paragraph{MAP inference with EM.} A general algorithm for finding MAP estimates in the presence of latent variables, is the Expectation Maximization (EM)~\cite{dempster1977em} algorithm. Since in our learning problem the expectation is intractable in general, approximations are required such as Monte Carlo EM~\cite{wei1990mcem} (MCEM) where an MCMC-based numerical expectation is used in the \emph{E}-step. The likelihood and prior are differentiable in the case under consideration, so gradients are easily obtained and Hybrid Monte Carlo~\cite{duane1987hmc} (HMC) can be used as a sampler, using $\pT(\bz | \bx) \propto p(\bx, \bz | \bT)$ for fixed $\bx$ and $\bT$. HMC has fast mixing properties in the high-dimensional and continuous case.
A faster method for approximate MAP inference is where we treat the latent variables $\bz$ as parameters, and we maximize $p(\bT, \bz | \bx) \propto p(\bx, \bz | \bT) p(\bT)$ w.r.t. $\bT$ and $\bz$ simultaniously using gradient ascent, which can be shown to optimize a lower bound of $p(\bT | \bx)$~\cite{neal1998em}. However, this comes with a relatively large risk of overfitting.

\paragraph{Full Bayesian inference with HMC.} In the case of full Bayesian estimation we are interested in estimating (or approximating) the full posterior distribution $p(\bT | \bx) \propto p(\bx | \bT) p(\bT)$. We can do this by sampling $\bT$ and $\bz$ simultanously from the joint density: $p(\bT, \bz | \bx) \propto p(\bx, \bz | \bT) p(\bT)$ and then keeping only samples of $\bT$. Similar to MCEM, HMC can be used to efficiently generate samples.

\subsection{Gradient-based information flow}\label{gradvalupdates}
The gradient-based approaches outlined above generally perform updates of $\bz$ using gradients of the joint PDF $\pT(\bx, \bz)$ in their inner loop, i.e. (1) for current values of $\bz$, acquire gradient information $\nabla_{\bz} \log \pT(\bx, \bz)$ by differentiation; (2) update $\bz$ using currently available gradient information. In Bayesian networks, the joint PDF $\pT(\bx, \bz)$ is available in factorized form (eq.~\eqref{eq:bnjoint}). From eq. \eqref{eq:bnjoint} it is easy to see that the gradient of the joint PDF with respect to some variable $X_j$ is only dependent on the variables that are in the the Markov blanket of $X_j$, because only the variables in the Markov blanket of $X_j$ share factors in which $X_j$ appear. The gradient of the joint log-PDF with respect to a variable is equal to the sum of gradients of the log of the connected factors in the factor graph.

In consequence, when performing a gradient-based value update of the variables (as outlined above), the current value of each variable only influences the new values of variables in its Markov blanket. The minimum number of factors that separate two variables in the factor graph determines the number of gradient steps required before information about the value of $X_i$ has reached $X_j$ and vice versa. We will show that it is possible to formulate the problem in an auxiliary form in which gradient-based inference and learning is generally faster.

\section{Inference and learning in auxiliary form}\label{sec:auxform}

We propose a method for transforming the original Bayesian network into an equivalent form in which gradient-based inference and learning is more efficient.

\subsection{Method}\label{method}

We will explain our method with a Bayesian network with $\bb{Z} \cup \bb{X} = \bb{V}$ where $\bb{X}$ is the set of observed variables and $\bb{Z}$ is the set of continuous latent variables that have parents in the graph. For the sake of clarity we don't include discrete latent variables in our explanation. Let the variables $X_j \in \bb{X}$ and $Z_j \in \bb{Z}$ be elements of these sets, with corresponding distributions $\pT(X_j|\bb{Pa}_j)$ and $\pT(Z_j|\bb{Pa}_j)$, where $\bT$ is again some global set of parameters. For each $j$, $\bb{Pa}_j$ is the set of parents of each variable; the parents of each node $j$ are subdivided into the sets $\bb{Pa}_j = \bb{X}_j \cup \bb{Z}_j$ where $\bb{X}_j \subseteq \bb{X}$ and $\bb{Z}_j \subseteq \bb{Z}$.
The Bayesian network models the joint PDF over the variables with factors $f_{X_j}(.)$ and $f_{Z_j}(.)$ as:
\begin{align}
\pT(\bb{X}, \bb{Z})
&= \prod_{X_j \in \bb{X}} \pT(X_j | \bb{Pa}_j)
\prod_{Z_j \in \bb{Z}} \pT(Z_j | \bb{Pa}_j) \\
&= \prod_{X_j \in \bb{X}} \pT(X_j | \bb{X}_j, \bb{Z}_j)
\prod_{Z_j \in \bb{Z}} \pT(Z_j | \bb{X}_j, \bb{Z}_j) \\
&= \prod_{X_j \in \bb{X}} f_{X_j}(X_j ; \bb{X}_j, \bb{Z}_j, \bT)
\prod_{Z_j \in \bb{Z}} f_{Z_j}(Z_j ; \bb{X}_j, \bb{Z}_j, \bT)
\end{align}

\begin{figure}[t]
\begin{center}
\begin{tabular}{cc}
\multicolumn{2}{c}{
\begin{tikzpicture}
\node[const] (z1) {$\cdots\,\,$};
\node[latent, right=of z1] (z2) {$Z_j$};
\node[const, right=of z2] (x) {$\,\,\cdots$};
\edge {z1} {z2};
\edge {z2} {x};

\node[const, right=2.0 of x] (_z1) {$\cdots\,\,$};
\node[det, right=of _z1] (_z2) {$\tZ_j$};
\node[latent, above=of _z2] (_e2) {$E_j$};
\node[const, right=of _z2] (_x) {$\,\,\cdots$};
\edge {_z1} {_z2};
\edge {_z2} {_x};
\edge {_e2} {_z2};
\end{tikzpicture}
} \\
\hspace{16 mm} (a) & \hspace{33 mm} (b) \\
\end{tabular}
\end{center}
\caption{
{\bf(a)} A continuous latent variable $Z_j$ with parents $\bb{Pa}_j$ and a conditional distribution $\pT(Z_j|\bb{Pa}_j)$. {\bf(b)} The auxiliary form where we replaced each $Z_j$ with $\tZ_j$ (with parents $\btPa_j$, where $\tZ = g_Z(\btPa_j, E_j, \bT)$, with auxiliary latent variable $E_j \sim \pT(E_j)$, such that $Z_j|\bb{Pa}_j$ equals $\tZ_j|\btPa_j$ in distribution. The diamond indicates a conditionally deterministic variable: the value of $\tZ_j$ is only deterministic when conditioned on both $\btPa_j$ and $E_j$.
}\label{auxformfig}
\end{figure}
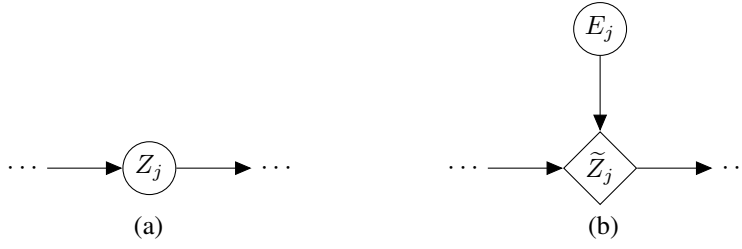

\paragraph{Step 1: Form an auxiliary Bayesian network.}
Form an auxiliary Bayesian network over variables $\bb{X}$, $\btZ$ and $\bb{E}$, where each original $Z_j \in \bb{Z}$ is replaced by a \emph{conditionally deterministic} variable (see eq.~\eqref{diracPDF}) $\tZ_j \in \btZ$, where each $\tZ_j$ has an extra parent $E_j \in \bb{E}$ with $E_j \sim \pT(E_j)$ (an auxiliary variable) that is a root node of the auxiliary network. Let in the auxiliary network $\btPa_j$ be each node's parents except any $E_j$, so $\btPa_j = \bb{X}_j \cup \bb{\tZ}_j$, and $\btpa_j$ a value of $\btPa_j$~\footnote{Latent variables that are root nodes of the graph can actually be copied to the auxiliary graph unchanged, since there is no computational advantage in putting them in auxiliary form, but for the sake of simplicity we will also convert them in this description.}. See figure~\ref{auxformfig}. 

\paragraph{Step 2: For each $\tZ_j \in \btZ$, define an appropriate generating function $g_j(.)$ and auxiliary latent variable $E_j \sim \pT(E_j)$. }
For each  $E_j$ we are going to define an appropriate distribution $E_j \sim \pT(E_j)$, and for each $\tZ_j$ an appropriate deterministic generating function $g_j(.)$ where:
\begin{align}\label{eq:auxtransform}
\tZ_j = g_j(\btPa_j, E_j, \bT)
\end{align}
such that $Z_j|\bb{Pa}_j$ and $\tZ_j|\btPa_j$ are equal in distribution:
\begin{align}
\forall \,\bpa_j, \bz_j:\hspace{5mm}
\pT(Z_j = \bz_j | \bb{Pa}_j = \bpa_j)
&= \pT(\tZ_j = \bz_j | \btPa = \bpa_j) 
\label{eq:auxcondition}
\end{align}
where:
\begin{align}
\pT(\tZ_j = \bz_j | \btPa = \bpa_j)  &= \int_{\beps_j} \pT(E_j = \beps_j) \pT(\tZ_j = \bz_j | \btPa = \bpa_j, E_j = \beps_j) \,d\beps_j \\
&= \Exp{\beps_j}{\pT(\tZ_j = \bz_j | \btPa = \bpa_j, E_j = \beps_j)}
\end{align}
In other words, for each continuous latent variable $Z_j \in \bb{Z}$ with parents $\bb{Pa}_j$ we define an auxiliary latent variable $E_j$ with an appropriate distribution $\pT(E_j)$, and a deterministic generating function $g_j(\bb{Pa}_j, E_j, \bT)$, such that the distributions of the new random variable $\tZ_j|\btPa_j$ and the original $Z_j|\bb{Pa}_j$ are equal. Note that the variable $\tZ_j|\btPa_j,E_j$ is deterministic, but $\tZ_j|\btPa_j$ is non-deterministic (when conditioned on only $\btPa_j$). Examples of valid choices of $E_j$ and $g_j(.)$ are discussed in section \ref{validauxforms}.

Under the condition of eq.~\eqref{eq:auxcondition} it can be shown that the joint $\bb{X}, \bb{Z}$ and the joint $\bb{X}, \bb{\tZ}$ are equal in distribution:
\begin{align}
\forall \,\bx, \bz :\hspace{3mm}
\pT(\bb{X} = \bx, \btZ = \bz)
&= \Exp{\beps}{\pT(\bb{X} = \bx, \btZ = \bz | \bb{E} = \beps)}
\\
&= \pT(\bb{X} = \bx, \bb{Z} = \bz)
\label{auxiliary-equality}
\end{align}

\paragraph{Step 3: Define the auxiliary PDF.}
Since each $E_j$ is a root node in the auxiliary network, its factor corresponds to its PDF we defined above: $\tf_{E_j}(E_j;\bT) = \pT(E_j)$. Each factor $\tf_{X_j}(.)$ corresponding to an observed variable $X_j$ in the auxiliary network is equal to the original factor $f_{X_j}(.)$, except that that any $Z_j$ in the function arguments is substituted for $\tZ_j$. Each factor $\tf_{\tZ_j}(.) = \pT(\btz_j|\btpa_j, \beps_j)$ corresponds to a conditionally deterministic variable $\tZ_j \in \btZ$ (see eq.~\eqref{diracPDF}). In effect the PDF of the joint distribution $\pT(\bb{X}, \bb{E}, \btZ)$ is zero almost everywhere. Interestingly, and this is important, the marginal PDF $\pT(\bb{X}, \bb{E})$ can be quite easily retrieved by marginalizing out the conditionally deterministic $\btZ$:
\begin{align}
\pT(\bx, \beps) &= \int_{\btz} \pT(\bx, \beps, \btz) \,d\btz\\
&= \int_{\btz}
\prod_{X_j \in \bb{X}} \pT(\bx_j|\btpa_j)
\prod_{\tZ_j \in \btZ} \pT(\btz_j|\btpa_j, \beps_j)
\prod_{E_j \in \bb{E}} \pT(\beps_j)  \,d\btz
\\
&= \prod_{X_j \in \bb{X}} \pT(\bx_j|\btpa_j)
\prod_{E_j \in \bb{E}} \pT(\beps_j)
\text{\hspace{5mm}where\hspace{5mm}$\forall \tZ_k \in \btZ : \btz_k = g_k(\btpa_k, \beps_k, \bT)$ } \\
&=
\prod_{X_j \in \bb{X}} \tf_{X_j}(\bx_j; \btpa_j, \bT)
\prod_{E_j \in \bb{E}} \tf_{E_j}(\beps_j; \bT)
\text{\hspace{5mm}where\hspace{5mm}$\forall \tZ_k \in \btZ : \btz_k = g_k(\btpa_k, \beps_k, \bT)$ }
\label{eq:auxpdf}
\end{align}
where the values of $\btz_k$ are computed (recursively) from their ancestors with $g_k(.)$. We call the marginalized PDF $\pT(\bx, \beps)$ of eq.~\eqref{eq:auxpdf} the \emph{auxiliary form} of the original problem.

\paragraph{Step 4: Fast inference and learning in auxiliary form.}
Important to note from eq.~\eqref{eq:auxpdf} is that factors (conditonal PDFs) $\tf_{X_j}(.)$ of observed variables $X_j$ with any $\tZ_k \in \btZ$ as their argument, can be expressed as being a function of $\tZ_k$'s parents, by substituting $\tZ_k$ by $g_k(.)$. If $\tZ_k$ then again has any other $\tZ_l$ as his parents, the factor $\tf_{X_j}(.)$ can again be expressed as a function of $\tZ_l$'s parents, and so on, recursively. As a result any factor $\tf_{X_j}(.)$ can be expressed as being a function with ancestral observed variables $X_k$'s and ancestral auxiliary variables $E_k$'s as its arguments.
Thus, the main advantage of the auxiliary form (eq.~\ref{eq:auxpdf}) is that the Markov blankets of the observed variables extend (recusively) through the conditionally deterministic variables, reaching (typically) much more variables. This leads to faster spread of information when performing gradient-based inference and learning, especially for Bayesian nets with many layers of interdependent continuous latent variables.

Evaluating and differentiating the auxiliary PDF for certain values of $\bx$ and $\beps$ is straightforward, which we explain in section \ref{implementation}. We can then e.g. apply gradient-based algorithms (section \ref{gradvalupdates}) to perform fast inference and learning.
Given equality~\ref{auxiliary-equality}, the values of $\btZ$, computed from corresponding values of $\bb{X}$ and $\bb{E}$, can at any point be treated as samples of $\bb{Z}$.

\subsection{Efficient Implementation}\label{implementation}

The method to efficiently evaluate and differentiate the auxiliary joint PDF given values $\bz$, $\beps$ of the variables $\bb{Z}$ and $\bb{E}$ is straightforward. First compute the corresponding values of each $\tZ_j$ according to their topological order. This order will make sure that for each $\tZ_j$, the value of its parents are computed first. Subsequently compute the values of all factors, and finally the value of the full joint. Computation of gradients of the joint w.r.t. variables and parameters can be done by the backpropagation algorithm~\cite{rumelhart1986backprop}. An alternative approach that requires substantially less manual differentiation work is through the use automatic differentiation software, such as Theano~\cite{bergstra2010theano}.

\subsection{Choice of $E_j$ and $g_j(.)$}\label{validauxforms}
An example of a valid choice of $g_j(.)$ and $E_j$ is based on the inverse transformation method (or Smirnov transform)~\cite{devroye86random}, a method for generating samples from a continuous target distribution by first sampling from a uniform distribution and then transforming it through the inverse CDF of the target distribution.
Indeed, an obvious choice would be to let $\tZ_j = g_j(.) = F^{-1}(E_j;\btPa_j,\bT)$, where $F^{-1}(E_j;\btPa_j,\bT)$ is the inverse CDF of $\pT(Z_j|\bb{Pa}_j)$ and where $E_j \sim \mathcal{U}(0,1)$.
For some distributions the inverse CDF is not available or not differentiable. In these cases the inverse CDF can often be accurately approximated using moderate-degree polynomials, e.g. used in software package R for generating samples from a Gaussian.

Besides the inverse CDF transform, there are often other valid options, e.g. in the case where $P(Z_j|\bb{Pa}_j) = \mathcal{N}(h_{\bT}(\bb{Pa}_j), \sigma_j^2)$, where $Z_j$'s distribution is univariate Gaussian with some variance $\sigma^2$ and a mean determined by its parents $\bb{Pa}_j$ through some (e.g. nonlinear) scalar function $h_{\bT}(.)$. In this case a valid choice of the generating function would be to let $E_j \sim \mathcal{N}(0, 1)$ and then let $g_j(.) = h_{\bT}(\btPa_j) + \sigma E_j$. This is a valid choice (eq.~\eqref{eq:auxcondition}) since:
\begin{align*}
\tz_j &= h_{\bT}(\btpa_j) + \sigma \epsilon_j \hspace{5mm}\text{ therefore }\hspace{5mm} \epsilon_j = (\tz_j - h_{\bT}(\btpa_j))/\sigma \eqnr \label{1234}\\
\pT(\tZ_j = z_j | \btPa = \bpa_j) &= \int_{\epsilon} \pT(E_j = \epsilon) \pT(\tZ_j = z_j | \btPa = \bpa_j, E_j = \epsilon) \,d\epsilon
& \\
&= \pT(E_j = (z_j - h_{\bT}(\bpa_j))/\sigma) 
 &\text{ (see eqs \ref{diracPDF} and \ref{1234})}\\
&= \mathcal{N}((z_j - h_{\bT}(\bpa_j))/\sigma; 0, 1)
= \mathcal{N}(h_{\bT}(\bpa_j); z_j, \sigma)
&\text{ (Gaussian PDF)} \\
&= \pT(Z_j=z_j|\bb{Pa}_j=\bpa_j) \eqnr\label{gaussianaux}
\end{align*}
A straightforward extension to the multivariate case (where $Z_i$ is a random vector) is where we treat each element as conditionally independent, in which case a valid solution is analogous to eq.~\eqref{gaussianaux}.

\subsection{Illustrative example}\label{sec:example}

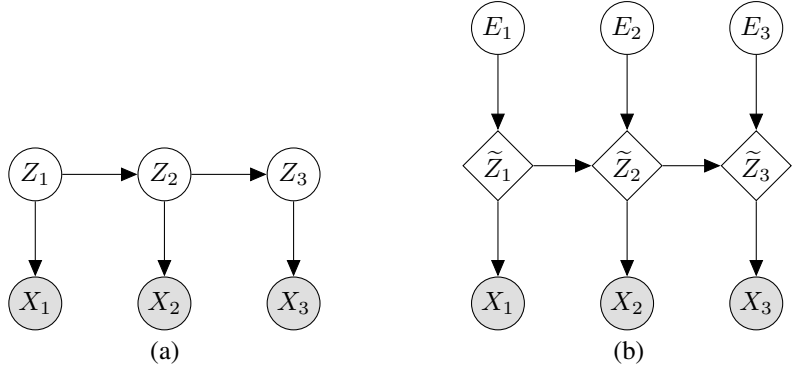
\begin{figure}[t]
\begin{center}
\begin{tabular}{cc}
\multicolumn{2}{c}{
\begin{tikzpicture}
\node[obs] (x1) {$X_1$};
\node[obs, right=of x1] (x2) {$X_2$};
\node[obs, right=of x2] (x3) {$X_3$};
\node[latent, above=of x1] (z1) {$Z_1$};
\node[latent, above=of x2] (z2) {$Z_2$};
\node[latent, above=of x3] (z3) {$Z_3$};
\edge {z1} {z2};
\edge {z1} {x1};
\edge {z2} {x2};
\edge {z2} {z3};
\edge {z3} {x3};

\node[obs, right=2.0 of x3] (_x1) {$X_1$};
\node[obs, right=of _x1] (_x2) {$X_2$};
\node[obs, right=of _x2] (_x3) {$X_3$};
\node[det, above=of _x1] (_z1) {$\tZ_1$};
\node[det, above=of _x2] (_z2) {$\tZ_2$};
\node[det, above=of _x3] (_z3) {$\tZ_3$};
\node[latent, above=of _z1] (_e1) {$E_1$};
\node[latent, above=of _z2] (_e2) {$E_2$};
\node[latent, above=of _z3] (_e3) {$E_3$};

\edge {_e1} {_z1};
\edge {_e2} {_z2};
\edge {_e3} {_z3};
\edge {_z1} {_z2};
\edge {_z2} {_z3};
\edge {_z1} {_x1};
\edge {_z2} {_x2};
\edge {_z3} {_x3};

\end{tikzpicture}
} \\
\hspace{18 mm} (a) & \hspace{33 mm} (b) \\
\end{tabular}
\end{center}
\caption{
{\bf(a)} A basic illustrative Bayesian network with three continuous latent variables and three observed variables, representing $\pT(X_1, X_2, X_3, Z_1, Z_2, Z_3)$.
{\bf(b)} The auxiliary form with conditionally deterministic variables $\tZ_1$, $\tZ_2$ and $\tZ_3$, chosen such that $\tZ_1 = g_1(E_1, \bT)$, $\tZ_2 = g_2(Z_1, E_2, \bT)$ and $\tZ_3 = g_3(\tZ_2, E_3, \bT)$, with auxiliary latent variables $E_2 \sim \pT(E_2)$ and $E_3 \sim \pT(E_3)$.
}\label{auxformexample}
\end{figure}

Take a simple Bayesian network as in figure~\ref{auxformexample} with three continuous latent variables $Z_1$, $Z_2$ and $Z_2$, and three observed variables $X_1$, $X_2$ and $X_3$. The joint PDF factorizes like
\begin{align*}
&\log \pT(\bx_1, \bx_2, \bx_3, \bz_1, \bz_2, \bz_3) \\
&= \log \pT(\bx_1|\bz_1)
+ \log \pT(\bx_2|\bz_2)
+ \log \pT(\bx_3|\bz_3)
+ \log \pT(\bz_3|\bz_2)
+ \log \pT(\bz_2|\bz_1)
+ \log \pT(\bz_1) 
\\
&= \log f_{X_1}(\bx_1;\bz_1,\bT)
+ \log f_{X_2}(\bx_2;\bz_2,\bT)
+ \log f_{X_3}(\bx_3;\bz_3,\bT)\\
&+ \log f_{Z_3}(\bz_3;\bz_2,\bT)
+ \log f_{Z_2}(\bz_2;\bz_1,\bT)
+ \log f_{Z_1}(\bz_1;\bT)
\eqnr\label{example-pdf-origform}
\end{align*}
Now imagine we are going to perform gradient-based inference as outlined in section \ref{algorithms}, in which we iteratively perform first-order gradient based updates of the latent variables given observations. The gradient of the joint PDF w.r.t. $\bz_1$ looks like:
\begin{align}
\nabla_{\bz_1} \log \pT(\bx_1, \bx_2, \bx_3, \bz_1, \bz_2, \bz_3)
= \nabla_{\bz_1} \log f_{X_1}(\bx_1;\bz_1,\bT)
+ \log f_{Z_1}(\bz_1;\bT)
\end{align}
Observe that $X_2$ and $X_3$ are not in $Z_1$'s Markov blanket: at each gradient step, the values of $\bz_1$ and $\bz_2$ influence each others new values, the values of $\bz_2$ and $\bz_3$ influence each other, and the values of $\bz_3$ and $\bx_3$ influence each other. It therefore takes three of such gradient steps for information about the value of $X_3$ to reach $Z_1$ in this example.

Now we are going to find an auxiliary form of the log-PDF in eq.~\eqref{example-pdf-origform}. As described in section~\ref{method}, we define an auxiliary network where $Z_1$, $Z_2$ and $Z_3$ are replaced by $\tZ_1$, $\tZ_2$ and $\tZ_3$, with an auxiliary latent variable $E_3$ and a function $g_3(\tZ_2,E_3,\bT)$ such that $Z_3|Z_2$ and $\tZ_3|\tZ_2$ equal in distribution. Similarly for $E_2$ and $E_1$. Note that putting $Z_1$ in auxiliary form is trivial and does not bring any advantages (since it doesn't have parents), but we'll do it anyway for the sake of consistency.
\begin{align}
\btz_1 = \beps_1
\text{\hspace{3mm}and\hspace{3mm}}
\btz_2 = g_2(\beps_2,\btz_1,\bT)
\text{\hspace{3mm}and\hspace{3mm}}
\btz_3 = g_3(\beps_3,\btz_2,\bT)
\end{align}
The auxiliary PDF (eq.\eqref{eq:auxpdf}) is:
\begin{align*}
\log \pT(\bx_1, \bx_2, \bx_3, \beps_1, \beps_2, \beps_3)
&= \log \tf_{X_1}(\bx_1;\btz_1,\bT)
+ \log \tf_{X_2}(\bx_2;\btz_2,\bT)
+ \log \tf_{X_3}(\bx_3;\btz_3,\bT)\\
&+ \log \tf_{E_1}(\beps_1;\bT)
+ \log \tf_{E_2}(\beps_2;\bT)
+ \log \tf_{E_3}(\beps_3;\bT)
\eqnr\end{align*}
Observe that each $\btz_j$ is a recursively a function of all $E_k$'s with $k \leq j$; therefore each variable $X_j$'s has all $E_k$'s with $k \leq j$ in its Markov blanket, while it only has $Z_j$ in its Markov blanket in the original network. This is also reflected in the gradients of the auxiliary joint PDF, e.g. w.r.t. $\beps_1$:
\begin{align*}
&\nabla_{\beps_1} \log \pT(\bx_1, \bx_2, \bx_3, \beps_1, \beps_2, \beps_3) = \eqnr\\ 
&\nabla_{\beps_1} \log \tf_{X_1}(\bx_1;\btz_1,\bT)
+ \nabla_{\beps_1} \log \tf_{X_2}(\bx_2;\btz_2,\bT)
+ \nabla_{\beps_1} \log \tf_{X_3}(\bx_3;\btz_3,\bT)
+ \nabla_{\beps_1} \log \tf_{E_1}(\beps_1;\bT)
\end{align*}


 

\section{Experiments}\label{sec:experiments}

Two experiments were performed to empirically evaluate the relative efficiency of the auxiliary form. 

\subsection{Generative model of MNIST digits}
In our first experiment we trained generative models of handwritten digits from the MNIST dataset~\cite{lecun1998gradient}, and compare convergence speed for inference in original versus auxiliary form. The first model consists of two layers of each continuous latent random vectors of size 64, and one layer with an observed binary random variable of size 768 (the digit images), connected like $Z_1 \to Z_2 \to X$. All variables are (random) vectors. The joint PDF is $\pT(\bx, \bz_1, \bz_2) = \pT(\bx|\bz_1) \pT(\bz_1|\bz_2) \pT(\bz_2)$, where $\pT(\bx|\bz_2) = \bx \bb{a} + (1-\bx)(1-\bb{a})$ where $\bb{a} = h(W_x \bz_2 + \bb{b}_x)$, where $h(.)$ is a piecewise sigmoid nonlinearity. The latent variables are connected with the PDF $\pT(\bz_2 | \bz_1) = \mathcal{N}(\bz_2; \tanh(W_{\bz_2} \bz_1 + \bb{b}_{\bz_2}), \bsigma_{\bz_2}^2 I)$, where $\mathcal{N}(.;.,.)$ is the Gaussian PDF. The PDF of the first layer is $\pT(\bz_1) = \mathcal{N}(\bz_1;0,\bsigma_{\bz_1}^2 I)$. The parameters are $\bT = \{W_{\bx}, \bb{b}_\bx, W_{\bz_2}, \bb{b}_{\bz_2}, \bsigma_{\bz_2}, \bsigma_{\bz_1} \}$. Note that the dependencies between layers bear similarities to a neural networks architecture. We also trained a network with a third layer of latent variables, like $Z_1 \to Z_2 \to Z_3 \to X$. 
The parameters follow a zero-centered Gaussian prior distribution: $\mathcal{N}(.;0,0.01 I)$, except for the $\sigma$'s which follow $\log \mathcal{N}(.;0,0.01 I)$. 

We used Monte Carlo EM (MCEM)~\cite{wei1990mcem} for learning the parameters of this generative network, where Hybrid Monte Carlo~\cite{duane1987hmc} was used for the E-step, and 5 steps of Adagrad~\cite{duchi2010adaptive} for the M-step. Each HMC iteration, five leapfrog steps were performed with a stepsize of 0.01. This setting worked well with an acceptance rate of around $50 \%$ for all experiments. Adagrad was used with an initial stepsize of $0.1$ for all experiments. Training was performed a random 10000 digits of the training set. All experiments were performed with the same initial parameters by sampling from the Gaussian $\bT \sim \mathcal{N}(0, 0.01)$. The auxiliary form of eq.~\ref{gaussianaux} was used.

\begin{figure}[t]
\centering
\begin{subfigure}[b]{0.57\textwidth}
\includegraphics[width=1\textwidth]{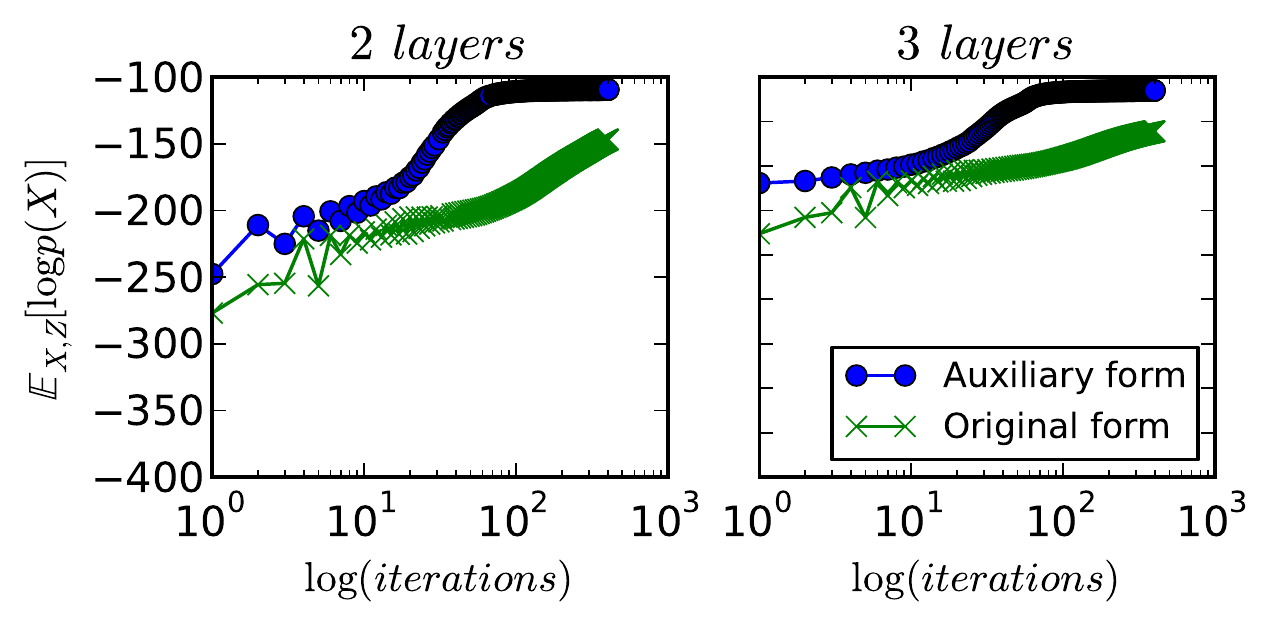}
\end{subfigure}
\begin{subfigure}[b]{0.32\textwidth}
\includegraphics[width=1\textwidth]{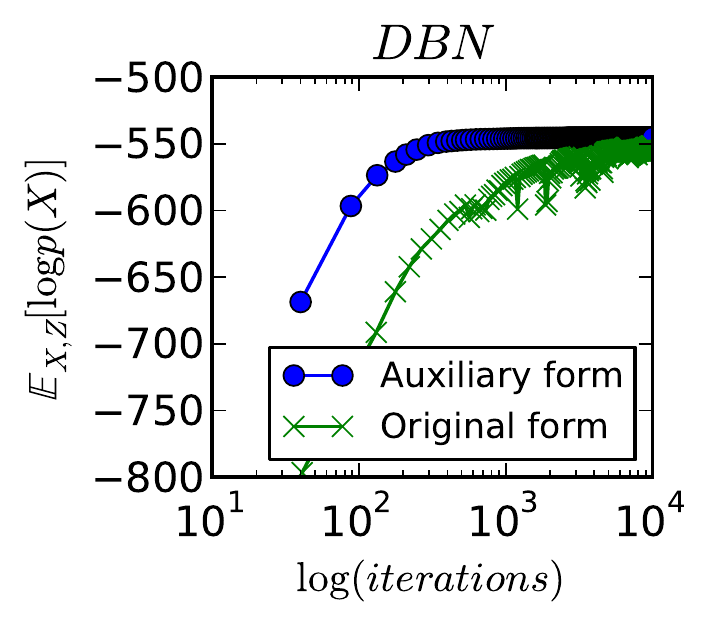}
\end{subfigure}
\caption{Left: convergence of log-likelihood for generative MNIST problem. Right: Convergence of log-likelihood for the dynamic Bayesian network (DBN).}
\label{fig:mnist}
\end{figure}

\paragraph{Results} MAP inference in original form progressed quite slowly, but much faster in auxiliary form, in both the 2-layer and 3-layer network, as expected from theory. Inference with auxiliary transform converged an order of magnitude faster in both cases. Interestingly, the 3-layer model did not improve upon the data likelihood. See figure \ref{fig:mnist}.

\subsection{Dynamical Bayesian Network}
In our second experiment we learned the parameters (also using MAP) of a dynamical Bayesian Network of length 10 and a structure as in section~\ref{sec:example}, with both latent variables and observed variables being continuous random vectors of size 10. 
Dependencies between latent variables were: $\pT(\bz_{t+1} | \bz_{t}) = \mathcal{N}(\bz_{t+1}; \tanh(W_{\bz} \bz_{t} + \bb{b}_{\bz}), \bsigma_{\bz}^2 I)$. Dependencies between latent variables and observed variables were $\pT(\bx_{t} | \bz_{t}) = \mathcal{N}(\bx_{t}; \tanh(W_{\bx} \bz_t + \bb{b}_{\bx}), \bsigma_{\bx}^2 I)$. Also, $\pT(\bz_0) = \mathcal{N}(0,\bsigma_{\bz_0}^2 I)$. The data was generated by sampling parameters from $\pT(\bz_0) = \mathcal{N}(0,I)$ and then forward sampling through the network, generating 100 datapointsm each with an assigment of observed variables for each timestep.
The parameter prior, initial parameters, the used auxiliary form and the MCEM hyperparameters were equal as in the MNIST experiment. 

\paragraph{Results} Similar to the MNIST experiment, MAP inference converged an order of magnitude faster in auxiliary form. See figure \ref{fig:mnist} (right).

\section{Conclusion}
The main contribution of this paper is that we show how to transform a Bayesian network with continuous latent variables to an auxiliary form. In this auxiliary form, latent variables are conditionally deterministic and can be integrated out. Instead, we sample from auxiliary variables. Since variables in the auxiliary form have larger Markov blankets, such inference should be generally faster. The method improves inference efficiency especially when there are multiple layers of dependent latent variables. The method is applicable to any conditional distribution with differentiable invertable CDFs, but we also show that easier transforms are possible as well. Efficiency is evaluated and confirmed empirically through measurement of inference efficiency with a generative model of the MNIST dataset, and a dynamic Bayesian network.

\newpage
\bibliography{bib}
\bibliographystyle{alpha} 

\end{document}

%% file: commands.tex
\usepackage{amsthm}
\usepackage{amsmath}
\usepackage{amsfonts}
\usepackage{amssymb}

\usepackage{tikz}
\usetikzlibrary{bayesnet}

\newcommand{\bb}[1]{\mathbf{#1}}
\newcommand{\bx}{\bb{x}}

\newcommand{\bT}{\boldsymbol{\theta}}
\newcommand{\by}{\bb{y}}
\newcommand{\bv}{\bb{v}}

\newcommand{\beps}{\boldsymbol{\epsilon}}
\newcommand{\bz}{\bb{z}}
\newcommand{\bpa}{\bb{pa}}

\newcommand{\pT}{p_{\bT}}
\newcommand{\bsigma}{\bb{\sigma}}
\newcommand{\tz}{\widetilde{z}}
\newcommand{\tZ}{\widetilde{Z}}
\newcommand{\tf}{\widetilde{f}}
\newcommand{\btZ}{\widetilde{\bb{Z}}}
\newcommand{\btz}{\widetilde{\bz}}

\newcommand{\btPa}{\widetilde{\bb{Pa}}}
\newcommand{\btpa}{\widetilde{\bb{pa}}}

\newcommand{\Exp}[2]{\mathbb{E}_{#1}\left[#2\right]}

\newcommand{\eqnr}{\addtocounter{equation}{1}\tag{\theequation}}

\theoremstyle{definition}

%% file: nips13_at.bbl
\newcommand{\etalchar}[1]{$^{#1}$}
\begin{thebibliography}{DKPR87}

\bibitem[BBB{\etalchar{+}}10]{bergstra2010theano}
James Bergstra, Olivier Breuleux, Fr{\'e}d{\'e}ric Bastien, Pascal Lamblin,
  Razvan Pascanu, Guillaume Desjardins, Joseph Turian, David Warde-Farley, and
  Yoshua Bengio.
\newblock Theano: a {CPU} and {GPU} math expression compiler.
\newblock In {\em Proceedings of the Python for Scientific Computing Conference
  (SciPy)}, volume~4, 2010.

\bibitem[CS05]{cobb2005nonlinear}
Barry~R Cobb and Prakash~P Shenoy.
\newblock Nonlinear deterministic relationships in {B}ayesian networks.
\newblock In {\em Symbolic and Quantitative Approaches to Reasoning with
  Uncertainty}, pages 27--38. Springer, 2005.

\bibitem[Dev86]{devroye86random}
Luc Devroye.
\newblock {\em Non-uniform Random Variate Generation}.
\newblock Springer, 1986.

\bibitem[DHS10]{duchi2010adaptive}
John Duchi, Elad Hazan, and Yoram Singer.
\newblock Adaptive subgradient methods for online learning and stochastic
  optimization.
\newblock {\em Journal of Machine Learning Research}, 12:2121--2159, 2010.

\bibitem[DKPR87]{duane1987hmc}
Simon Duane, Anthony~D Kennedy, Brian~J Pendleton, and Duncan Roweth.
\newblock Hybrid {M}onte {C}arlo.
\newblock {\em Physics letters B}, 195(2):216--222, 1987.

\bibitem[DLR77]{dempster1977em}
Arthur~P Dempster, Nan~M Laird, and Donald~B Rubin.
\newblock Maximum likelihood from incomplete data via the {EM} algorithm.
\newblock {\em Journal of the Royal Statistical Society. Series B
  (Methodological)}, pages 1--38, 1977.

\bibitem[LBBH98]{lecun1998gradient}
Yann LeCun, L{\'e}on Bottou, Yoshua Bengio, and Patrick Haffner.
\newblock Gradient-based learning applied to document recognition.
\newblock {\em Proceedings of the IEEE}, 86(11):2278--2324, 1998.

\bibitem[Min01]{minka2001expectation}
Thomas~P Minka.
\newblock Expectation propagation for approximate bayesian inference.
\newblock In {\em Proceedings of the Seventeenth conference on Uncertainty in
  artificial intelligence}, pages 362--369. Morgan Kaufmann Publishers Inc.,
  2001.

\bibitem[Mur02]{murphy2002dynamic}
Kevin~Patrick Murphy.
\newblock {\em Dynamic {Bayesian} networks: representation, inference and
  learning}.
\newblock PhD thesis, University of California, 2002.

\bibitem[Nea93]{neal1993probabilistic}
Radford~M Neal.
\newblock Probabilistic inference using markov chain monte carlo methods.
\newblock 1993.

\bibitem[NH98]{neal1998em}
Radford~M Neal and Geoffrey~E Hinton.
\newblock A view of the {EM} algorithm that justifies incremental, sparse, and
  other variants.
\newblock In {\em Learning in graphical models}, pages 355--368. Springer,
  1998.

\bibitem[Pea82]{pearl1982reverend}
Judea Pearl.
\newblock {\em Reverend Bayes on inference engines: A distributed hierarchical
  approach}.
\newblock Cognitive Systems Laboratory, School of Engineering and Applied
  Science, University of California, Los Angeles, 1982.

\bibitem[RHW86]{rumelhart1986backprop}
David~E Rumelhart, Geoffrey~E Hinton, and Ronald~J Williams.
\newblock Learning representations by back-propagating errors.
\newblock {\em Nature}, 323(6088):533--536, 1986.

\bibitem[WT90]{wei1990mcem}
Greg~CG Wei and Martin~A Tanner.
\newblock A {M}onte {C}arlo implementation of the {EM} algorithm and the poor
  man's data augmentation algorithms.
\newblock {\em Journal of the American Statistical Association},
  85(411):699--704, 1990.

\end{thebibliography}
